\documentclass[10pt,twocolumn,letterpaper]{article}
\usepackage[accsupp]{axessibility}

\usepackage[pagenumbers]{wacv} 

\usepackage{graphicx}
\usepackage{amsmath}
\usepackage{amssymb}
\usepackage{booktabs}
\newcommand{\bm}[1]{{\mbox{\boldmath $#1$}}}

\usepackage{algorithmic}
\usepackage{algorithm}

\usepackage[pagebackref,breaklinks,colorlinks]{hyperref}

\usepackage[capitalize]{cleveref}
\crefname{section}{Sec.}{Secs.}
\Crefname{section}{Section}{Sections}
\Crefname{table}{Table}{Tables}
\crefname{table}{Tab.}{Tabs.}

\def\wacvPaperID{1975} 
\def\confName{WACV}
\def\confYear{2024}

\begin{document}

\title{Controlling Rate, Distortion, and Realism: Towards a Single Comprehensive Neural Image Compression Model}

\author{Shoma Iwai\hspace{8mm} Tomo Miyazaki\hspace{8mm} Shinichiro Omachi\\
Graduate School of Engineering, Tohoku University, Japan\\
{\tt\small shoma.iwai.b4@dc.tohoku.ac.jp, \{tomo, shinichiro.omachi.b5\}@tohoku.ac.jp}
}
\maketitle

\begin{abstract}
   In recent years, neural network-driven image compression (NIC) has gained significant attention. Some works adopt deep generative models such as GANs and diffusion models to enhance perceptual quality (realism). A critical obstacle of these generative NIC methods is that each model is optimized for a single bit rate. Consequently, multiple models are required to compress images to different bit rates, which is impractical for real-world applications. To tackle this issue, we propose a variable-rate generative NIC model. Specifically, we explore several discriminator designs tailored for the variable-rate approach and introduce a novel adversarial loss. Moreover, by incorporating the newly proposed multi-realism technique, our method allows the users to adjust the bit rate, distortion, and realism with a single model, achieving ultra-controllability. Unlike existing variable-rate generative NIC models, our method matches or surpasses the performance of state-of-the-art single-rate generative NIC models while covering a wide range of bit rates using just one model.
   Code will be available at \url{https://github.com/iwa-shi/CRDR}.
\end{abstract}

\section{Introduction}

\label{sec:intro}
Image compression is a fundamental technique for efficient image storage and transmission. In recent years, neural-network-based image compression (NIC) methods have received much attention\cite{Balle2017, Balle2018, He_2022_CVPR, Liu_2023_CVPR}. Most NIC models are optimized to minimize the rate-distortion loss function, represented as $R+\lambda D$. $R$ is the bit rate after compression, $D$ is the distortion between the original image and its compressed counterpart, typically measured by mean squared error (MSE), and $\lambda$ determines the balance between distortion and bit rate. State-of-the-art NIC models\cite{Liu_2023_CVPR, He_2022_CVPR} have surpassed the rate-distortion performance of the latest standard codec, Versatile Video Coding (VVC)\cite{Bross2021vvc}.

\begin{figure}[t]
    \centering
    \includegraphics[width=\linewidth]{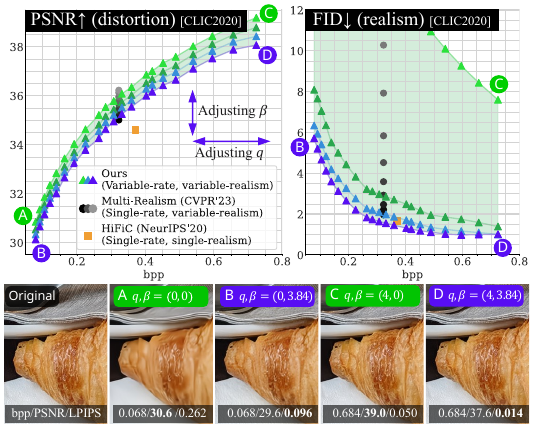}
    \caption{Top left: rate-distortion (measured by PSNR) and top right: rate-realism (measured by FID) performance using a \textbf{single model}. ``bpp" stands for bits-per-pixel. While state-of-the-art GAN-based NIC methods, \textit{Multi-Realism}\cite{Agustsson2023} (capable of adjusting distortion-realism trade-off) and \textit{HiFiC}\cite{Mentzer2020} are optimized to a single bit rate, our method can control the balance between rate, distortion, and realism, covering the green area with just one model. This is achieved by adjusting two input parameters, $q$ and $\beta$, which control the rate and the distortion-realism trade-off\cite{Blau2018}, respectively. Bottom: the original image and compression results of our method. It illustrates that our method can handle different compression settings like \textcircled{\scriptsize A} low-rate and low-distortion mode and \textcircled{\scriptsize D} high-rate and high-realism mode.}
    \label{fig:thumbnail}
\end{figure}

General NIC models have two drawbacks. First, one NIC model is optimized for a specific bit rate, requiring multiple distinct models to compress an image at various bit rates. Second, NIC models trained with the rate-distortion loss function tend to produce blurred images, particularly at lower bit rates, resulting in low human perceptual quality.

For the first problem, several variable-rate NIC models have been studied\cite{Sun_2021_ACMMM, Song2021, Choi2019, Cui2021}, which can compress images at diverse bit rates using just one model. SOTA variable-rate NIC models\cite{Sun_2021_ACMMM, Cai2022acmmm} demonstrate comparable performance as single-rate counterparts.

For the second problem, some works\cite{Mentzer2020, Agustsson2023, theis2022, ghouse2023} have employed generative models such as generative adversarial networks (GAN)\cite{Goodfellow2014} and denoising diffusion probabilistic models (DDPM)\cite{ho2020denoising} to enhance perceptual quality (realism) of reconstructions. Such generative NIC models can produce realistic images even at low bit rates.

However, only a few studies have worked on the variable-rate generative NIC model\cite{Gao2021_CLIC, Ma_2021_CLIC, ghouse2023, Gupta2022a}, and they come with certain limitations. Specifically, \cite{Gao2021_CLIC, Ma_2021_CLIC} offers only a narrow range of adjustable bit rates, while \cite{ghouse2023, Gupta2022a} suffers from poor performance compared to SOTA single-rate generative NIC models.

In this study, we propose a novel variable-rate GAN-based NIC model. Our approach differs from existing variable-rate generative NIC models in three ways.
Firstly, we explore various discriminator designs tailored for the variable-rate GAN-based NIC model. 
Our comparison shows that the discriminator design has a substantial impact on performance.
Secondly, we propose a novel adversarial loss function termed as HRRGAN (Higher Rate Relativistic GAN) to stabilize training.
Thirdly, we adopt a beta-conditioning\cite{Agustsson2023} to control the distortion-realism trade-off. Consequently, our method can adjust the rate-distortion-realism trade-off\cite{Blau2019} using a single model, achieving ultra-controllability as shown in Fig~\ref{fig:thumbnail}.

Our contributions are summarized as follows:

\noindent \textbf{GAN-based training tailored for variable-rate NIC model}:
We offer a comparison and analysis of the various discriminator designs for the variable-rate GAN-based NIC model. Moreover, we introduce a novel adversarial loss function.

\noindent \textbf{High controllability and high performance}:
    Our model can adjust the balance between rate, distortion, and realism within a single NIC model.
    Even with this high controllability, our model matches or outperforms the performance of state-of-the-art single-rate generative NIC models\cite{Mentzer2020, Agustsson2023} on quantitative evaluation. To the best of our knowledge, this is the first variable-rate model that achieves the same or better performance as SOTA single-rate generative NIC models.

\begin{figure*}[h]
    \centering
    \includegraphics[width=.85\linewidth]{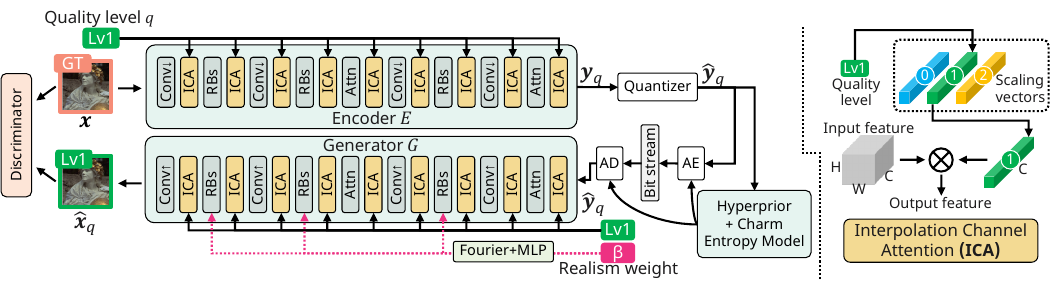}
    \caption{The overview of our NIC model. \textit{RBs} and \textit{Attn} in the encoder and generator stands for residual blocks and attention module used in ELIC\cite{He_2022_CVPR}. \textit{ICA} is an interpolation channel attention layer\cite{Sun_2021_ACMMM} (see right side for detail). \textit{AE} and \textit{AD} represent an arithmetic encoder, and arithmetic decoder, respectively. For the generator, we use beta-conditioning\cite{Agustsson2023} to control the realism of the reconstruction.}
    \label{fig:overview}
\end{figure*}

\section{Related Works}

\subsection{Single Rate Neural Image Compression (NIC)}
Since a VAE-based NIC model proposed by Balle et al.\cite{Balle2017}, a number of NIC models have been studied\cite{Balle2017, Balle2018, Minnen2020, He_2022_CVPR}. Typical NIC models consist of an encoder, decoder, and entropy model.  Most studies focus on entropy modeling schemes to improve compression performance, such as hyperprior\cite{Balle2018}, context model\cite{Minnen2018, Lee2019}, transformer-based context model\cite{Burakhan_eccv_2022, mentzer2023m2t, qian2021entroformer}, channel autoregressive model (Charm)\cite{Minnen2020}, and checkerboard context model\cite{He_2021_CVPR}. For the encoder and decoder, attention layers\cite{Cheng2020, Chen2021}, swin-transformer-based\cite{Liu_2021_ICCV} architecture\cite{zhu2022transformerbased, Zou_2022_CVPR}, and CNN-Transformer mixed architecture\cite{Liu_2023_CVPR} have been proposed. State-of-the-art NIC models\cite{He_2022_CVPR, Liu_2023_CVPR, mentzer2023m2t} outperform the latest coding standard, Versatile Video Coding (VVC)\cite{Bross2021vvc}.

\subsection{Variable Rate NIC}
In real-world scenarios, users adjust the compression ratio based on storage capacity or internet latency. However, most NIC models are optimized for a single rate point.
Therefore, multiple models are required to compress images at different bit rates, increasing training and model storage costs. To address this issue, several variable-rate NIC methods have been studied\cite{Choi2019, Cui2021, Sun_2021_ACMMM, Song2021, Yang2020spl, Chen2020icassp, Baldassarre2023icassp, gao2022flexible, Cai2022acmmm}. These methods take an additional input representing the target quality and adjust the bit rate accordingly. Choi et al.\cite{Choi2019} have introduced a conditional convolution to implement a variable-rate model.  Cui et al.\cite{Cui2021} proposed a gain-unit to modify the quantization step and control the information lost. Sun et al.\cite{Sun_2021_ACMMM} proposed an interpolation channel attention (ICA), allowing fine rate control without sacrificing compression performance.

\subsection{Generative NIC}
To improve the perceptual quality (realism) of compressed images, GANs\cite{Goodfellow2014} and diffusion model\cite{ho2020denoising} have been incorporated into NIC. Since the pioneer work\cite{Rippel2017}, several efforts have been made to improve the performance and training stability of GAN-based NIC\cite{Mentzer2020, Agustsson2023, el-nouby2023image}. Agustsson et al.\cite{Agustsson2023} have introduced a conditional generator to control the distortion-realism trade-off\cite{Blau2018} within a single model. Specifically, users can decide between reconstructions that have high PSNR values but appear blurry and those that look more realistic but have a lower PSNR. More recently, some diffusion-based NIC models have been proposed\cite{ghouse2023, hoogeboom2023highfidelity, yang2023lossy}. While these methods suffer from slow inference due to the iterative process, they achieve impressive perceptual quality.

However, most of the generative NIC models are single-rate models, which is not practical in real-world applications. There are only a few existing variable-rate generative NIC models that cover a wide range of bit rates.\cite{Gupta2022a, ghouse2023}.  Gupta et al.\cite{Gupta2022a} utilize a spatial importance map to realize a variable-rate model. Another study by Ghouse et al.\cite{ghouse2023} has proposed a diffusion-based NIC model, DIRAC. It leverages a pre-trained variable-rate NIC model and enhances the reconstruction quality with a diffusion model.  However, both methods are inferior to state-of-the-art single-rate models\cite{Agustsson2023, Mentzer2020} in terms of PSNR and FID. Moreover, DIRAC\cite{ghouse2023} requires a computationally expensive diffusion process to generate high-quality images. In this work, we bridge the performance gap between the variable-rate generative NIC model and SOTA single-rate models\cite{Mentzer2020, Agustsson2023, hoogeboom2023highfidelity} by analyzing the discriminator design and introducing a novel adversarial loss function.

\section{Proposed Method}
\subsection{NIC pipeline}
\label{sec:nic_pipeline}
We begin by outlining the entire pipeline of our variable-rate NIC model in Fig~\ref{fig:overview}.
To control the rate-distortion-realism trade-off with one model, our model takes three inputs: the original image $\bm{x}$, quality level $q\in \{0, 1, \cdots, Q-1\}$, and realism weight $\beta \in [0, \beta_{max}]$. Higher $q$ results in a higher rate, and higher $\beta$ results in higher realism. $q$ and $\beta$ are sampled randomly during training.
The model is based on the state-of-the-art single-rate GAN-based NIC model, Multi-realism\cite{Agustsson2023}, which incorporates ELIC\cite{He_2022_CVPR} encoder, channel autoregressive model (Charm\cite{Minnen2020}) and the beta-conditional generator\cite{Agustsson2023}.
We insert interpolation channel attention (ICA) layers\cite{Sun_2021_ACMMM} into the encoder and generator to make the model variable-rate. Each ICA layer (Fig~\ref{fig:overview} right) has $Q$ learnable scaling vectors, and one of them is selected and applied according to $q$.

The compression process is as follows. First, the encoder $E$ extracts latent representation $\bm{y}_q=E(\bm{x}, q)$. The latent $\bm{y}_q$ is then quantized into $\hat{\bm{y}}_q$ using scalar quantization. Since the quantization process is not differentiable, we use straight-through estimation (STE) during training as used in \cite{Minnen2020}.  The quantized code $\hat{\bm{y}}_q$ is transformed into a bitstream losslessly by the arithmetic encoder with the probability distribution $p(\hat{\bm{y}}_q)$ estimated by the entropy model. For the details on entropy estimation, please refer to the original Charm paper\cite{Minnen2020}. The quality level $q$ is also included in the bit stream, which occupies less than 1 byte. On the decoder side, the bitstream is first decoded back into the quantized code $\hat{\bm{y}}_q$ and quality level $q$ with the arithmetic decoder. The generator $G$ then reconstructs an image $\hat{\bm{x}}_q=G(\hat{\bm{y}}_q, q, \beta)$. As in \cite{Agustsson2023}, $\beta$ is first embedded with Fourier encoding\cite{mildenhall2020nerf} and MLPs and injected into the residual blocks.  Since we use GAN-based training, a discriminator is applied during the training. We describe the details of the discriminator architecture in the next section.

\begin{figure*}[h]
    \centering
    \includegraphics[width=.8\linewidth]{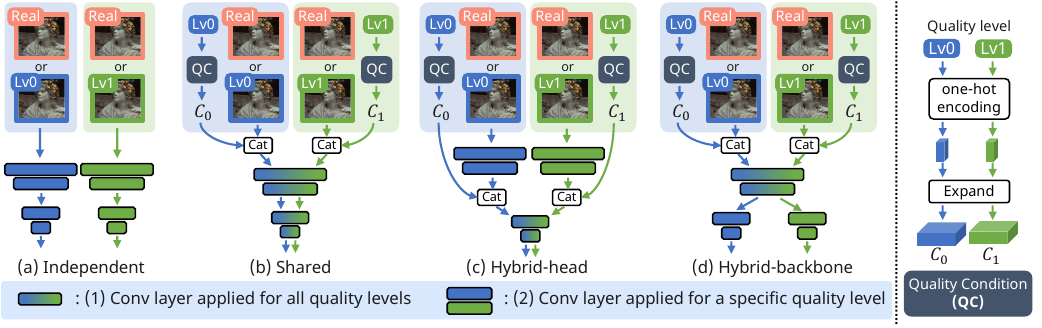}
    \caption{The discriminator designs that we consider. Discriminators take an original image or its reconstruction as input and estimate the reality of the input. They consist of two kind of convolution layers: (1) layers applied for all quality levels and (2) layers applied for a specific quality level. For layer (1), we introduce a quality condition (right).}
    \label{fig:discriminator}
\end{figure*}

\subsection{Exploring Discriminator}
\label{sec:disc}
In this section, we discuss and explore various discriminator designs for variable-rate NIC model. The discriminator is trained to distinguish real (original) and fake (reconstructed) images, and the NIC model learns to reconstruct images indistinguishable from the discriminator. Although the discriminator plays an important role in GAN-based training, its design remains underexplored in current variable-rate GAN-based NIC methods\cite{Gao2021_CLIC, Ma_2021_CLIC, Gupta2022a}.

Unlike single-rate GAN-based NIC models, variable-rate NIC models produce varying-quality images according to the quality level $q$. Since $q$ is chosen randomly at every training step, the reconstruction quality can significantly vary between training steps. This behavior differs from single-rate GAN-based methods, which motivates us to design a discriminator architecture specifically tailored for the variable-rate NIC.

Fig~\ref{fig:discriminator} shows the various discriminator designs we explored. The base architecture of all discriminators is a CNN patch-discriminator\cite{Isola2017} as used in\cite{Agustsson2023, Mentzer2020}. The output dimension is $\frac{H}{16} \times \frac{W}{16}$, where $H, W$ are the height and width of the input image, respectively.  These discriminators have two types of convolution layers: (1) those applied for all quality levels and (2) those applied for a specific quality level.  For layer (1), we introduce a quality condition to provide the discriminator information about the quality level. Specifically, given $q$, we employ one-hot encoding and expand this to get a conditional feature $C \in \{0,1\}^{h\times w \times Q}$. $C$ is concatenated with the input of the discriminator's first convolution layer (1). While layer (1) can capture features common across all quality levels (inherent in the NIC model), layer (2) can learn unique features on the individual quality level. We aim to find which features are more beneficial for the NIC model by comparing different designs. We describe the details of each design as follows.

\textbf{(a) Independent discriminator} consists of $Q$ distinct sub-discriminators.
Each sub-discriminator is applied to reconstructions of one particular quality level. However, it cannot leverage features common across various quality levels.

\textbf{(b) Shared discriminator} is a single CNN discriminator applied across all quality levels. While it can capture the features ubiquitous across all quality levels, it cannot learn the quality-level-specific features.

\textbf{(c), (d) Hybrid-head and -backbone discriminator} have both types of layers to leverage quality-level-specific and common features.  (c) has independent backbones and a shared prediction head, whereas (d) has a shared backbone and independent prediction heads. By comparing them, we aim to analyze the results when each layer type is applied to low- and high-level features.

Based on the comparison of performance (see Fig~\ref{fig:ablation_d}), we use \textit{(a) independent discriminator}. We will discuss the detailed results in Sec.\ref{sec:exp_disc}. Note that the discriminator is used only in the training; thus, using different layers for each quality level does not affect the inference process, such as model size and encoding/decoding speed. 

\begin{figure}[t]
    \centering
    \includegraphics[width=.9\linewidth]{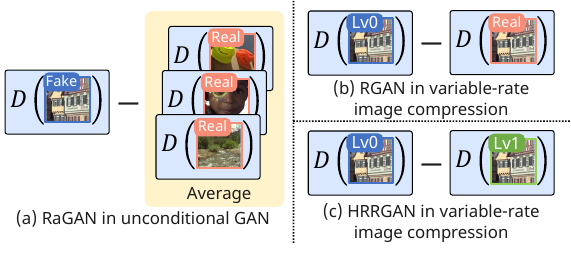}
    \caption{How to calculate the relative ``reality score" on (a) RaGAN in unconditional GAN, (b) RGAN in variable-rate image compression, and (c) HRRGAN in variable-rate image compression.}
    \label{fig:hrrgan}
\end{figure}

\begin{figure*}[h]
    \centering
    \includegraphics[width=.9\linewidth]{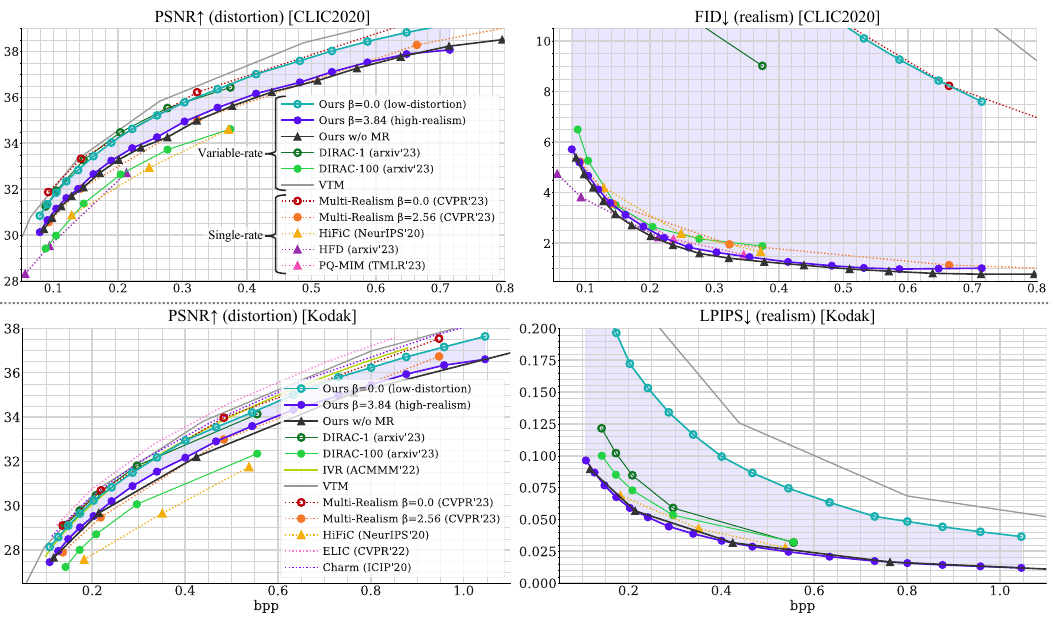}
    \caption{Quantitative results on CLIC2020 test (top) and Kodak dataset (bottom). We use PSNR to evaluate the rate-distortion performance and FID and LPIPS for the rate-realism performance. Solid lines represent variable-rate methods, while dashed lines denote single-rate methods. As for the markers, circles ($\bullet$ and $\circ$) represent variable-realism NIC, triangles ($\blacktriangle$) indicate generative NIC, and lines without markers indicate non-generative methods. We report LPIPS on CLIC2020 dataset in the supplementary material.
}
    \label{fig:quant_result}
\end{figure*}

\subsection{Higher Rate Relativistic GAN (HRRGAN)}
\label{sec:hrrgan}
In this section, we describe our novel adversarial loss function, Higher Rate Relativistic GAN (HRRGAN), which is inspired by Relativistic average GAN (RaGAN)\cite{jolicoeur-martineau2018}. In some generative NIC methods\cite{Gao2021_CLIC, Cheng_2021_CLIC}, RaGAN has been used to improve perceptual quality. In RaGAN, the adversarial losses for the generator and discriminator are calculated as follows:
\begin{align}
\label{eq:ragan}
    p_{r}(x_r, x_f) = \sigma (D(x_r) - \mathbb{E}_{x_f} [D(x_f)])\nonumber \\
    p_{f}(x_r, x_f) = \sigma (D(x_f) - \mathbb{E}_{x_r} [D(x_r)])\nonumber\\
    \mathcal{L}_{\text{RaGAN}}^{G} = -\log p_{f}(x_r, x_f) - \log(1- p_{r}(x_r, x_f))\\
    \mathcal{L}_{\text{RaGAN}}^{D} = -\log p_{r}(x_r, x_f) - \log(1- p_{f}(x_r, x_f)),
\end{align}
where $D$ is the discriminator, and $x_r, x_f$ are real and fake images, respectively. $\sigma(\cdot)$ and $\mathbb{E}_{x_f}[\cdot]$ represent sigmoid function and the average operation for all $x_f$ in the mini batch, respectively. Intuitively, in RaGAN, the discriminator $D$ estimates the ``reality score" of an input image instead of the probability that the input image is real as the standard GAN\cite{Goodfellow2014}. Then, the generator is trained so that the reality score of the fake image are higher than those of real images on average.

Our HRRGAN is different from RaGAN in two ways. First, we omit the average computation in Eq~\ref{eq:ragan}, which is the same loss function as the Relativistic standard GAN (RGAN)\cite{jolicoeur-martineau2018}.
In unconditional GAN, the average computation is necessary because $x_r$ and $x_f$ are not aligned as shown in Fig~\ref{fig:hrrgan}(a). However, in the image compression task, the fake image is a reconstructed version of the original, ensuring that the $x_r$ and $x_f$ are spatially aligned (Fig~\ref{fig:hrrgan}(b)). To leverage this alignment and calculate the relative reality score for each region, we do not average the discriminator's outputs as follows:
\begin{align}
    \mathcal{L}_{\text{RGAN}}^{G} = -\log \sigma (D(x_f) - D(x_r))\\
    \label{eq:rgan} \mathcal{L}_{\text{RGAN}}^{D} = -\log \sigma (D(x_r) - D(x_f)).
\end{align}

Second, we replace the real image with a reconstruction of higher quality when training the NIC model in Eq~\ref{eq:rgan}. In RGAN, the compression model is trained so that the reconstruction is estimated as more realistic than the original. We found that this approach imposes an over-penalty on the NIC model, resulting in excessive loss values even for successful reconstructions. To mitigate this problem, when the model reconstructs image $\bm{\hat{x}}_q$ with a quality level $q$, we use another reconstruction $\bm{\hat{x}}_{q+1}$ with quality level $q+1$ to calculate the relative reality score. The loss function of our HRRGAN is represented as follows:
\begin{align}
\label{eq:hrrgan}
    \mathcal{L}_{\text{HRRGAN}}^{G} = -\log \sigma(D(\hat{\bm{x}}_{q}) - \text{sg} (D(\hat{\bm{x}}_{q+1}))) \\
    \mathcal{L}_{\text{HRRGAN}}^{D} = -\log \sigma (D(\bm{x}) - D(\hat{\bm{x}}_{q})),
\end{align}
where $\rm{sg}$ denotes the stop gradient operation, preventing the NIC model from generating less realistic $\hat{\bm{x}}_{q+1}$ to minimize Eq~\ref{eq:hrrgan}.
Although $\bm{\hat{x}}_{q+1}$ has higher quality than $\bm{\hat{x}}_{q}$, its quality is not high as $\bm{x}$. As a result, HRRGAN relaxes the loss function for the NIC model and reduces the chance of over-penalizing, leading to more balanced training.
Note that the same loss function as RGAN is used for the discriminator.

\subsection{Training}
\label{sec:training}
Following \cite{Mentzer2020}, our training consists of two stages. In the first stage, we train the model without adversarial loss using the following loss function:
\begin{align}
\label{eq:loss_first}
    \mathcal{L}_{1st} = \lambda^{(q)}_R R(\hat{\bm{y}}_q) + \lambda_d d(\bm{x}, \hat{\bm{x}}_q) + \mathcal{L}_P(\bm{x}, \hat{\bm{x}}_q),
\end{align}
where $R, d, \mathcal{L}_P$ represent the bit rate estimated by the entropy model, MSE, and LPIPS\cite{Zhang2018lpips}, respectively. The weight of the bit rate $\lambda^{(q)}_R\in \{\lambda^{(0)}_R, \lambda^{(1)}_R, \cdots, \lambda^{(Q-1)}_R \}$ is selected according to the quality level.

In the second stage, we fine-tune the model with the proposed adversarial loss as follows:
\begin{align}
\label{eq:loss_second}
    \mathcal{L}_{2nd} = \lambda^{(q)}_R R(\hat{\bm{y}}_q) + \lambda_d d(\bm{x}, \hat{\bm{x}}_q) \nonumber \\
    + \beta(\lambda_P \mathcal{L}_P(\bm{x}, \hat{\bm{x}}_q) + \lambda_{\text{adv}} \mathcal{L}_{\text{HRRGAN}}^{G}),
\end{align}
where $\beta$ balances the influence of LPIPS and adversarial loss. As a result, the NIC model learns to generate high-realism images with a higher value of input $\beta$.

\begin{figure*}[t]
    \centering
    \includegraphics[width=.95\linewidth]{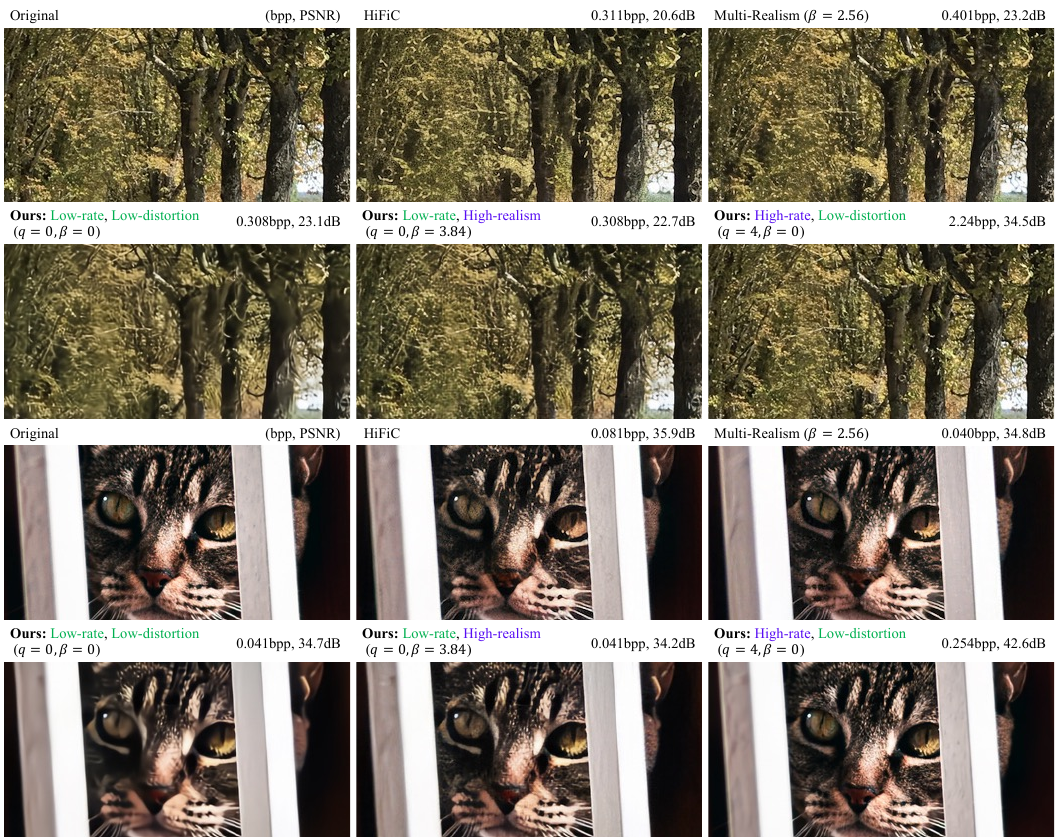}
    \caption{Comparing the original image to reconstructions of our model and state-of-the-art generative NIC models, HiFiC\cite{Mentzer2020} and Multi-Realism\cite{Agustsson2023} on CLIC2020 dataset. We show our three reconstructions with different configurations: $(q, \beta)=(0, 0), (0, 3.84), (4, 0)$.} 
    \label{fig:qualitative}
\end{figure*}

\section{Experiments}
\subsection{Experimental settings}
\noindent \textbf{Dataset.}
In our experiment, we trained our model on the subset of OpenImage dataset\cite{OpenImages2}, which contains about 1M images. We randomly crop the images into $256 \times 256$ patches and apply random horizontal flipping. The batch size is set to 8.  We evaluate models on standard benchmarks for the image compression task, Kodak\cite{kodakDataset} (24 images) and CLIC2020 test dataset\cite{clic2020website} (428 images).

\noindent \textbf{Evaluation.}
We use PSNR for distortion (fidelity) performance and Frech\'{e}t Inception Distance (FID)\cite{Heusel2017FID} and Learned Perceptual Image Patch Similarity (LPIPS)\cite{Zhang2018lpips} for realism (perceptual quality) performance. We followed the protocol in \cite{Mentzer2020} to calculate FID. We did not use FID for Kodak because it contains only 24 images.

\begin{figure*}[t]
    \centering
    \includegraphics[width=.9\linewidth]{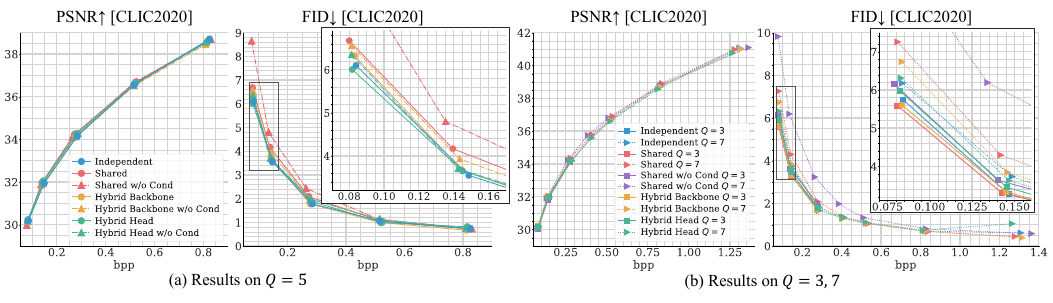}
    \caption{Quantitative comparison among different discriminator designs. \textit{w/o Cond} indicates that the discriminator does not take the condition $C$ as input. (a) shows the results on $Q=5$. (b) shows the results on $Q=3$ and $Q=7$}
    \label{fig:ablation_d}
\end{figure*}

\noindent \textbf{Training settings.}
We optimize the model using Adam\cite{Kingma2014}. We set the training steps of each training stage to $2M$ and $3M$. In each stage, the initial learning rate is set to $1.0 \times 10^{-4}$ and decayed to $1.0 \times 10^{-5}$ for the last 20\% of the total iterations.  The same learning rate settings are applied to the discriminator. The number of quality levels is $Q=5$, and $\beta_{max}$ is set to $5.12$ as in \cite{Agustsson2023}. For the loss function, we set $\{\lambda_R^{(0)}, \cdots, \lambda_R^{(4)}\}=\{3.4, 1.3, 0.4, 0.12, 0.05\}$, $\lambda_d=150$, $\lambda_P=2/\beta_{max}$, and $\lambda_\text{adv}=0.002/\beta_{max}$.

\noindent \textbf{Baselines.}
We compare our variable-rate GAN-based NIC model with existing compression methods. These methods are divided into three groups: generative NIC models, non-generative NIC models, and non-learning-based codec. Generative NIC models include GAN-based methods (Multi-realism\cite{Agustsson2023}, HiFiC\cite{Mentzer2020}, and PQ-MIM\cite{el-nouby2023image}) and diffusion-based methods (HFD\cite{hoogeboom2023highfidelity} and DIRAC\cite{ghouse2023}). Non-generative NIC models, which are optimized for PSNR, include ELIC\cite{He_2022_CVPR}, Charm\cite{Minnen2020}, and IVR\cite{Sun_2021_ACMMM}. As a non-learning-based codec, we use the latest codec, VTM\cite{VTM171} (software based on VVC\cite{Bross2021vvc}). Except for DIRAC, IVR, and VTM, all baselines are single-rate methods, requiring separate models for different bit rates. Multi-realism\cite{Agustsson2023} and DIRAC\cite{ghouse2023} can control the distortion-realism trade-off with one model. Specifically, \textit{Multi-realism $\beta=0$} and \textit{DIRAC-1} represent low-distortion mode, while \textit{Multi-realism $\beta=2.56$} and \textit{DIRAC-100} are high-realism mode.

\subsection{Quantitative Comparison}
Figure~\ref{fig:quant_result} illustrates quantitative results on CLIC2020 and Kodak datasets at different bit rates. Although our model is trained on $Q$ quality levels, we perform fine rate-tuning during inference by interpolating the scaling vectors\cite{Sun_2021_ACMMM}, obtaining reconstructions on 17 different rate points.  For \textit{Ours}, we show results on low-distortion mode ($\beta=0$) and high-realism mode ($\beta=3.84$). We report results on other $\beta$ in the supplementary material. \textit{Ours w/o MR} is the model trained with the fixed realism weight, $\beta=2.56$.

First, we compare our high-realism mode (\textit{Ours ($\beta=3.84$)}) with other generative NIC models. On CLIC2020 dataset, \textit{Ours ($\beta=3.84$)} surpassed the variable-rate generative model, DIRAC-100, on both PSNR and FID. Compared to the SOTA single-rate model, Multi-Realism ($\beta=2.56$)\cite{Agustsson2023}, our method demonstrates superior FID and competitive PSNR, despite using a single model for different rates. Although the recent diffusion-based method, HFD\cite{hoogeboom2023highfidelity}, achieves the best FID at low rate points, we significantly excel in PSNR. On Kodak dataset, we outperform DIRAC-100\cite{ghouse2023} and HiFiC\cite{Mentzer2020} on both PSNR and LPIPS. Furthermore, on both datasets, \textit{Ours ($\beta=3.84$)} achieves comparable performance as \textit{Ours w/o MR}, indicating that the $\beta$-conditioning\cite{Agustsson2023} does not hurt the realism performance even in variable-rate model.

Our low-distortion mode ($\beta=0.0$) achieves comparable performance as other low-distortion mode models, \textit{Multi-Realism ($\beta=0.0$)} and \textit{DIRAC-1}, in terms of PSNR on CLIC2020. Moreover, the PSNR values of \textit{Ours ($\beta=0.0$)} on Kodak match other distortion-oriented variable-rate models (\textit{DIRAC-1} and \textit{IVR}).
These results suggest that both modes of our model achieve strong performance while covering a wide range of rates.

\subsection{Qualitative Comparison}
We present qualitative comparisons with state-of-the-art generative NIC models\cite{Mentzer2020, Agustsson2023} in Fig~\ref{fig:qualitative}. In the first example, our $(q,\beta)=(0,3.84)$ reconstruction preserves the texture of leaves more effectively than HiFiC\cite{Mentzer2020}. Furthermore, it has comparable perceptual quality as Multi-Realism\cite{Agustsson2023}. In the second sample, while HiFiC,  Multi-Realism, and our $(q,\beta)=(0,3.84)$ mode display similar levels of realism, HiFiC uses a double bit rate. Comparing our three reconstructions, our $(q,\beta)=(0,0)$ reconstructions appear blurry but have higher PSNR than $(q,\beta)=(0,3.84)$ counterpart, confirming that our model controls the distortion-realism trade-off. Moreover, although our $(q,\beta)=(4,0)$ reconstructions use high bit rates, they preserve contents faithfully and achieve higher PSNR. In conclusion, the qualitative results demonstrate that our method works well for different use cases with only one model while matching the visual performance of state-of-the-art methods.

\subsection{Impact of Discriminator Design}
\label{sec:exp_disc}
In this section, we analyze the impact of discriminator designs.
We used the NIC models without Charm\cite{Minnen2020} and trained them with fixed $\beta=2.56$ to save computation costs.

Fig~\ref{fig:ablation_d}(a) compares the rate-distortion-realism performance on different designs on $Q=5$. \textit{w/o Cond} indicates that a quality condition $C$ is not used.
Through the comparison, we make the following observations.
First, the discriminator designs have a significant impact on FID, especially at low bit rates. However, there is no significant difference in terms of PSNR.
Second, the information about the quality level is crucial. In Fig~\ref{fig:ablation_d}(a), only the \textit{Shared w/o Cond} takes no information about the quality level (i.e., no level-specific layer and quality level condition) and results in clearly the worst FID.
Third, the level-specific layer is beneficial. \textit{Shared} does not have a level-specific layer and performs worse than other designs with level-specific layers, suggesting the necessity of the quality-level-specific layers.
Finally, convolution layers applied for all quality levels are not necessary. While \textit{Independent} does not have a layer shared across all levels, the performance is comparable to or better than hybrid designs: \textit{Hybrid Backbone} and \textit{Hybrid Head}. It indicates that capturing features common across all quality levels may not benefit the NIC model, whether at shallow or deep layers.

To further examine the impact of discriminators, we trained the models with wider and narrower bit rate ranges. Specifically, we present the results on $Q=7$ (roughly 0.08 $\sim$ 1.3bpp) and $Q=3$ (roughly 0.08 $\sim$ 0.3bpp) in Fig~\ref{fig:ablation_d}(b). For $Q=7$, we observed a similar trend to Fig~\ref{fig:ablation_d}(a), where \textit{Shared} and \textit{Shared w/o Cond} result in clearly high FID. On the other hand, for $Q=3$, \textit{Shared} performs comparably as \textit{Independent}, and the performance gap between \textit{Shared w/o Cond} and other designs is less substantial than in $Q=5,7$. These results indicate that the discriminator design is particularly crucial for the NIC model with a wider bit rate range.
In addition, Fig~\ref{fig:ablation_d}(b) demonstartes that \textit{Independent} performs robustly on different $Q$.

\begin{figure}[t]
    \centering
    \includegraphics[width=.9\linewidth]{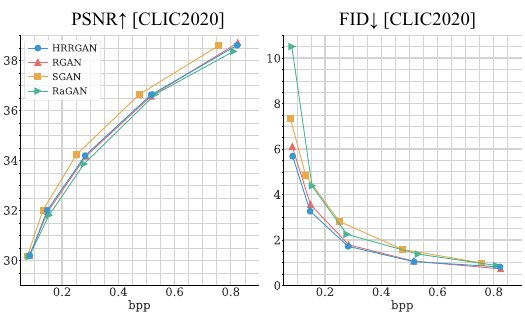}
    \caption{Results on different adversarial loss functions: HRRGAN (ours), Relativistic GAN (RGAN), standard GAN (SGAN)\cite{Goodfellow2014}, and Relativistic average GAN (RaGAN)\cite{jolicoeur-martineau2018}.}
    \label{fig:ablation_hrrgan}
\end{figure}

\begin{figure}[t]
    \centering
    \includegraphics[width=\linewidth]{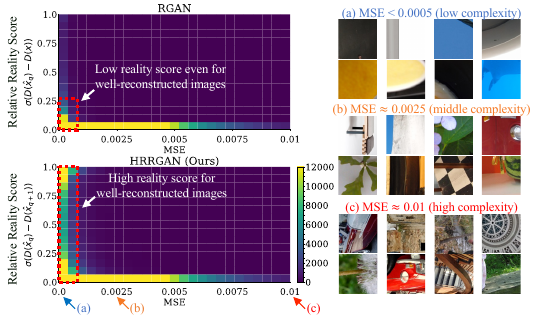}
    \caption{The two-dimensional histogram shows the relationship between reconstruction MSE and relative ``reality score" estimated by the discriminator on RGAN and HRRGAN. The lower the reality score is, the higher the adversarial loss becomes. The right side shows examples of training samples with different MSE values. It suggests that the samples with low MSE (a) have lower complexity than those with higher MSE (b) and (c).}
    \label{fig:analysis_hrrgan}
\end{figure}

\subsection{Effect of HRRGAN}
To verify the effectiveness of our HRRGAN, we compare it with other adversarial loss functions: standard GAN (SGAN)\cite{Goodfellow2014}, Relativistic GAN (RGAN), and Relativistic average GAN (RaGAN)\cite{jolicoeur-martineau2018}. We used NIC models without Charm\cite{Minnen2020} and trained them with fixed $\beta=2.56$ to save computation costs. Fig~\ref{fig:ablation_hrrgan} shows the results on CLIC2020. Regarding FID, HRRGAN performs best, particularly at lower bit rates. As discussed in Sec~\ref{sec:hrrgan}, HRRGAN relaxes the loss function of RGAN, leading to more balanced training and better performance. Although SGAN achieves the highest PSNR at middle and high rates, it performs worse on FID. Moreover, RGAN consistently surpasses RaGAN, indicating that the average calculation of RaGAN harms performance in the image compression task.

To further analyze the effect of HRRGAN, we show a 2D histogram representing the relationship between the reconstruction MSE and relative ``reality score" of RGAN and HRRGAN in Fig~\ref{fig:analysis_hrrgan}. For this analysis, we generated 4000 reconstructions with dimensions of $256\times 256$ from OpenImage validation dataset. These reconstructions are fed into the trained discriminator. Then, we calculate MSE and the relative reality score: $D(\hat{\bm{x}}_q)-D(\bm{x})$ for RGAN, and $D(\hat{\bm{x}}_q)-D(\hat{\bm{x}}_{q+1})$ for HRRGAN.
The histogram shows that RGAN assigns low reality scores (i.e., high adversarial loss) even to samples with extremely low MSE like Fig~\ref{fig:analysis_hrrgan} (a). This leads to an excessive penalty for easy and well-reconstructed samples. In contrast, HRRGAN tends to output high reality scores (i.e., low adversarial loss) for low MSE samples. It mitigates the risk of over-penalty and encourages the NIC model to focus on refining complex and challenging samples like Fig~\ref{fig:analysis_hrrgan} (b) and (c).

\section{Conclusion}
We have proposed a novel NIC model that can control the rate-distortion-realism trade-off\cite{Blau2019} with one model by adjusting two input parameters, $q$ and $\beta$. We have tried various discriminator designs and found that quality-level-specific layers are important for variable-rate generative NIC. Moreover, inspired by RaGAN\cite{jolicoeur-martineau2018}, we proposed HRRGAN to avoid over-penalty. In the experiments, our method achieved state-of-the-art rate-distortion-realism performance with one NIC model.

For limitations, although our method realizes high controllability, $q$ and $\beta$ control rate and realism uniformly. Consequently, it cannot perform pixel-level control. In real-world applications, however, precisely preserving specific regions (e.g., small faces) is important.  Therefore, integrating pixel-level control as in \cite{Song2021, gao2022flexible} will be our future work.

\noindent \textbf{Acknowledgements.} This work was supported by JST SPRING, Grant Number JPMJSP2114. We thank Zhengmi Tang for his valuable feedback on this paper.

{\small
\bibliographystyle{ieee_fullname}
\bibliography{references}
}


\appendix

\section{Detailed Discriminator Architecture}
\begin{figure}[h]
    \centering
    \includegraphics[width=.9\linewidth]{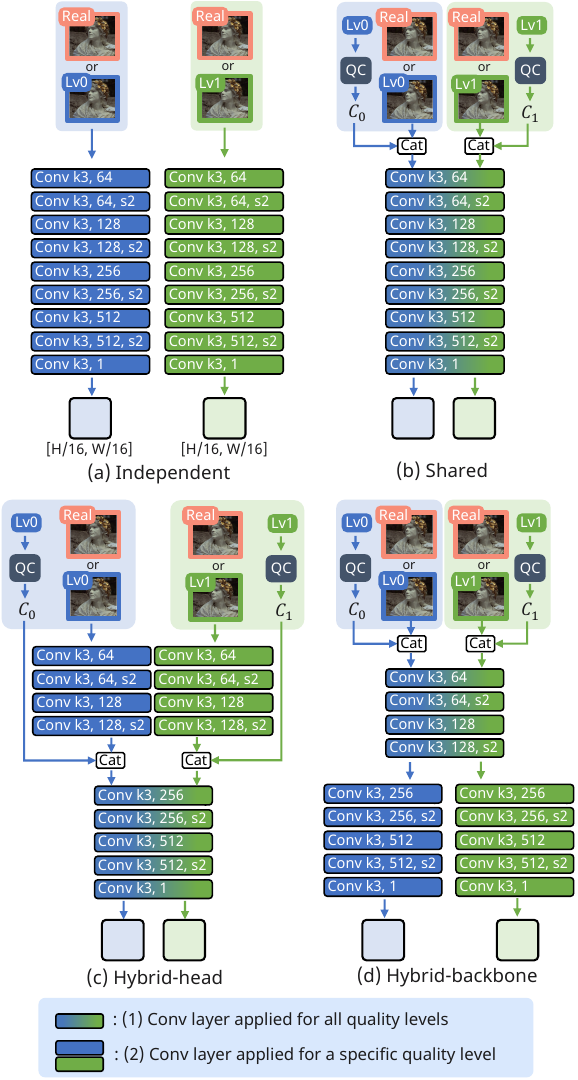}
    \caption{The detail of the discriminator architecture. 
    Numbers within the convolution layers represent the kernel size, output channel, and stride, respectively. For instance, "Conv k3, 256, s2" denotes a convolution layer with a kernel size of 3, an output channel of 256, and a stride of 2. Each discriminator's output dimensions are $(H/16, W/16)$, where $H$ and $W$ are the height and width of the image, respectively.
}
    \label{fig:sup_d_detail}
\end{figure}

We show the detailed architectures of each discriminator design in Fig~\ref{fig:sup_d_detail}. We apply leaky ReLU after each convolution layer except in the final layer.

\section{Algorithm of HRRGAN}
Algorithm~\ref{alg1} shows the procedure for calculating the HRRGAN loss function. In the algorithm, $\text{NIC}$, $D$, and $\text{sg}$ indicate the NIC model, discriminator, and stop-gradient operation, respectively. As shown in Algorithm~\ref{alg1}, we use the original image $\bm{x}$ to calculate $p_r$ when $q=Q-1$ because the NIC model cannot reconstruct an image with $q=Q$.

\begin{figure}[!t]
\begin{algorithm}[H]
    \caption{HRRGAN loss function}
    \label{alg1}
    \begin{algorithmic}[1]
    \STATE \textbf{Input:} Original image $\bm{x}$
    \STATE Uniformly sample realism weight $\beta \in [0, \beta_{max}]$
    \STATE Uniformly sample quality level $q \in \{0, 1, \cdots, Q-1\}$
    \STATE $\hat{\bm{x}}_q \leftarrow \text{NIC}(\bm{x}, q, \beta)$ 
    \IF{$q < Q - 1$}
    \STATE $\hat{\bm{x}}_{q+1} \leftarrow \text{NIC}(\bm{x}, q+1, \beta)$ 
    \STATE $p_r \leftarrow \text{sigmoid}(D(\hat{\bm{x}}_q) - \text{sg}(D(\hat{\bm{x}}_{q+1})))$
    \ELSE
    \STATE $p_r \leftarrow \text{sigmoid}(D(\hat{\bm{x}}_q) - \text{sg}(D(\bm{x})))$
    \ENDIF
    \STATE $p_f \leftarrow \text{sigmoid}(D(\bm{x}) - D(\hat{\bm{x}}_q))$
    \STATE $\mathcal{L}_{HRRGAN}^{G} \leftarrow -\log p_r$
    \STATE $\mathcal{L}_{HRRGAN}^{D} \leftarrow -\log p_f$
    \end{algorithmic}
\end{algorithm}
\end{figure}

\section{Model Size}
Table~\ref{tab:param} shows the number of parameters in our encoder and generator.
As described in Sec~3.1, we adopt the encoder and generator in ELIC\cite{He_2022_CVPR} as a base architecture and incorporate Interpolation Channel Attention (ICA) layers\cite{Sun_2021_ACMMM} and $\beta$-conditioning\cite{Agustsson2023} to adjust rate and distortion-realism trade-off, respectively. ICA layers are used in the encoder and generator, while $\beta$-conditioning is used only in the generator.
As shown in Table~\ref{tab:param}, using these two modules results in an approximate parameter increase of $2.7M$. It demonstrates that our approach significantly saves model storage costs compared to employing separate NIC models optimized for distinct rates and distortion-realism balances.

\begin{table}[t!]
\begin{tabular}{@{}lcccc@{}}
\toprule
            & ICA        & $\beta$-cond & Encoder                                                   & Generator                                                    \\ \midrule
Base        &            &              & 7.34M                                                     & 10.72M                                                     \\ \midrule
Ours w/o MR & \checkmark &              & \begin{tabular}[c]{@{}c@{}}7.36M\\ (+0.019M)\end{tabular} & \begin{tabular}[c]{@{}c@{}}10.74M\\ (+0.024M)\end{tabular} \\ \midrule
Ours full   & \checkmark & \checkmark   & \begin{tabular}[c]{@{}c@{}}7.36M\\ (+0.019M)\end{tabular} & \begin{tabular}[c]{@{}c@{}}13.38M\\ (+2.66M)\end{tabular}  \\ \bottomrule
\end{tabular}
\label{tab:param}
\caption{Parameter counts for the encoder and generator across various configurations. ``ICA" stands for interpolation channel attention\cite{Sun_2021_ACMMM}, while ``$\beta$-cond" refers to $\beta$-conditioning\cite{Agustsson2023}. \textit{Base} represents the encoder and decoder (generator) used in ELIC\cite{He_2022_CVPR}. The numbers in parentheses represent the increase in parameters compared to \textit{Base}.}
\label{tab:param}
\end{table}

\section{Additional Results}

\subsection{Results on different realism weights}
\begin{figure*}[t]
    \centering
    \includegraphics[width=\linewidth]{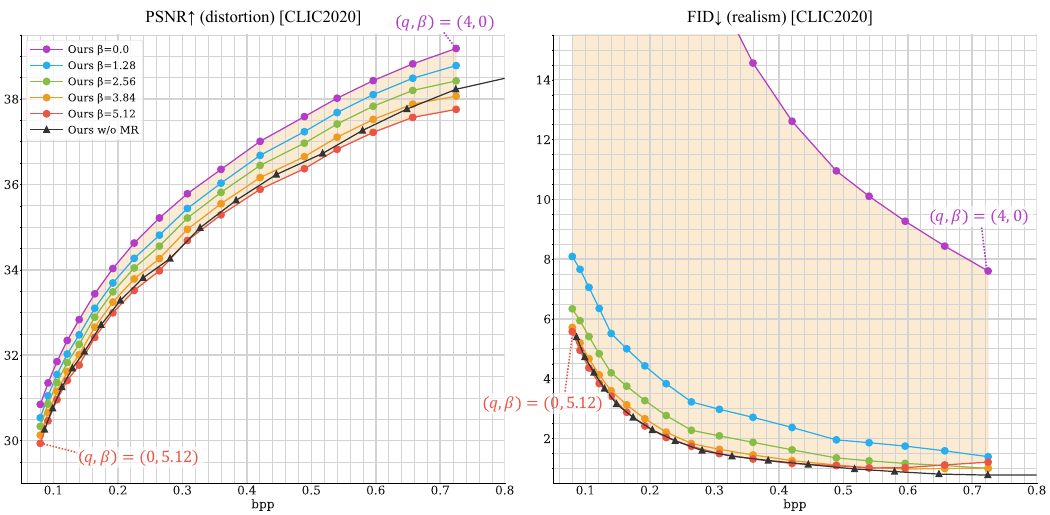}
    \caption{
    Quantitative comparison of different input realism weights $\beta$ on CLIC2020 test dataset. \textit{Ours w/o MR} represents a baseline model trained with fixed $\beta=2.56$. These results demonstrate that our model effectively balances the distortion-realism trade-off by adjusting input $\beta$.}
    \label{fig:sup_clic_beta}
\end{figure*}

We show the quantitative results on different realism weights $\beta =\{0, 1.28, 2.56, 3.84, 5.12\}$ in Fig~\ref{fig:sup_clic_beta}.
These results illustrate that different $\beta$ results in different balances between distortion and realism.
Specifically, smaller $\beta$ achieves high PSNR, indicating higher pixel-level fidelity. On the other hand, larger $\beta$ yields lower FID, indicating high realism.
Although $\beta=3.84$ consistently surpasses $\beta=5.12$ in terms of PSNR, the difference in FID between $\beta=3.84$ and $\beta=5.12$ is marginal. Based on this observation, we selected $\beta=3.84$ as our high-realism mode in our main experiments.

\subsection{LPIPS evaluation on CLIC2020 dataset}
\begin{figure*}[t]
    \centering
    \includegraphics[width=0.7\linewidth]{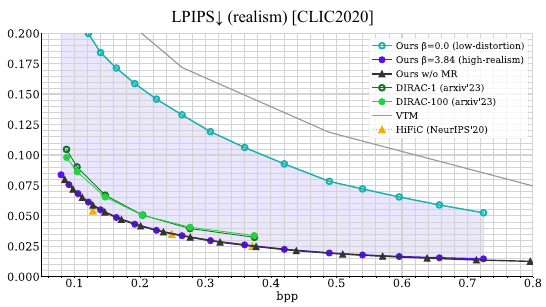}
    \caption{LPIPS\cite{Zhang2018lpips} results on CLIC2020 test dataset.}
    \label{fig:sup_clic_lpips}
\end{figure*}

Fig~\ref{fig:sup_clic_lpips} shows the results of LPIPS\cite{Zhang2018lpips} on CLIC2020 datset.
Our high-realism mode ($\beta=3.84$) matches the performance of HiFiC\cite{Mentzer2020} (single-rate model) and outperforms DIRAC\cite{ghouse2023} (variable-rate model).

\begin{figure*}[t]
    \centering
    \includegraphics[width=\linewidth]{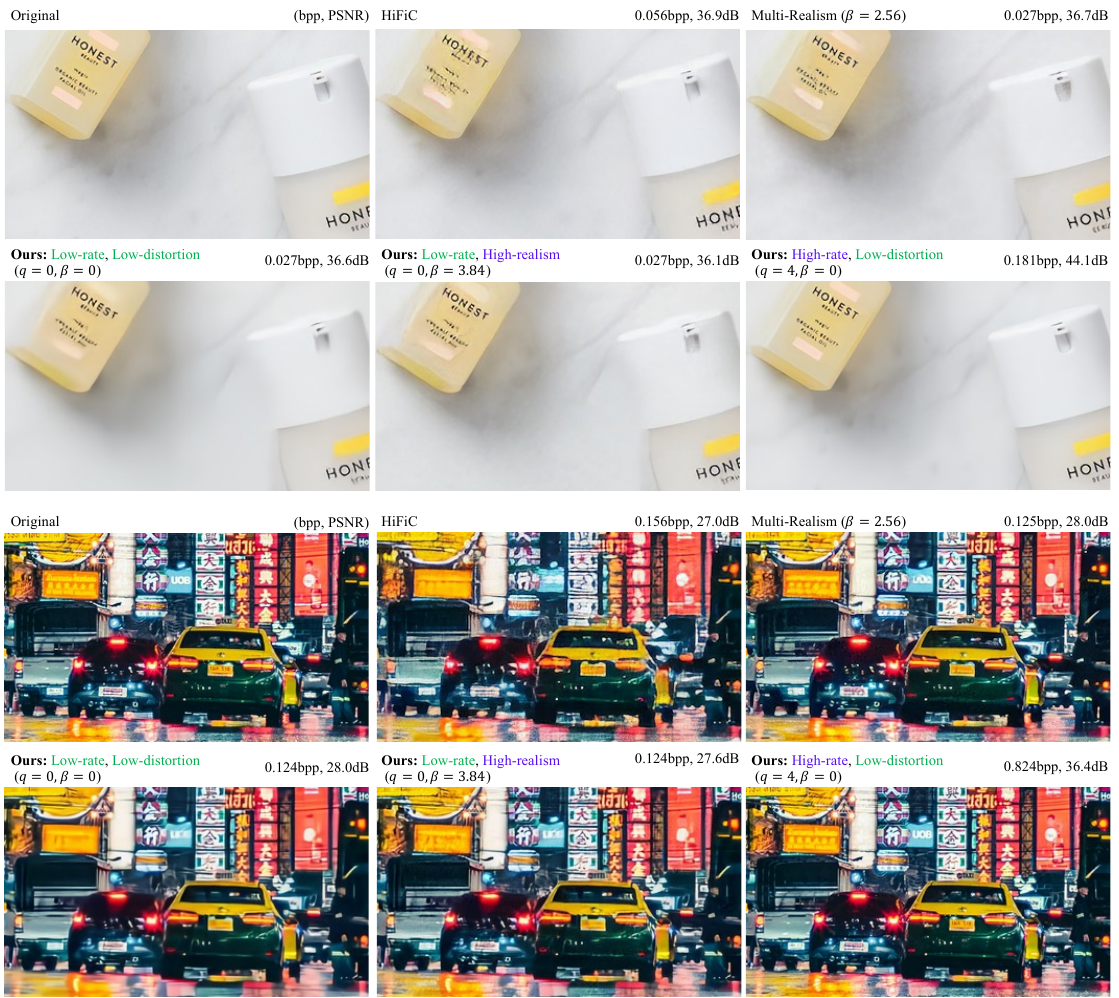}
    \caption{Qualitative comparison on CLIC2020 dataset.}
    \label{fig:sup_clic_qualitative_1}
\end{figure*}

\begin{figure*}[t]
    \centering
    \includegraphics[width=\linewidth]{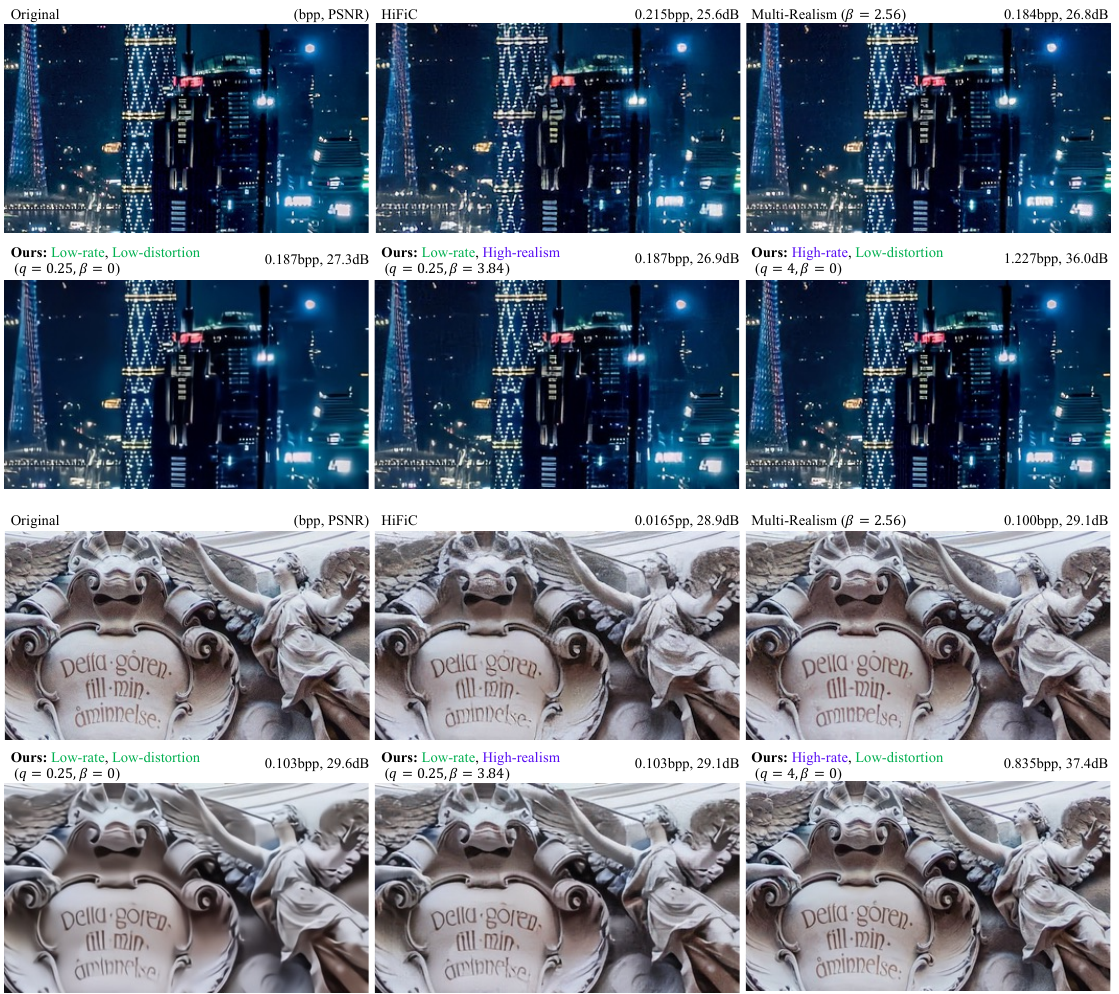}
    \caption{Qualitative comparison on CLIC2020 dataset. In \textit{Ours}, non-integer $q$ indicates that we interpolated the scaling vectors for fine rate control as in\cite{Sun_2021_ACMMM}.}
    \label{fig:sup_clic_qualitative_2}
\end{figure*}

\begin{figure*}[t]
    \centering
    \includegraphics[width=\linewidth]{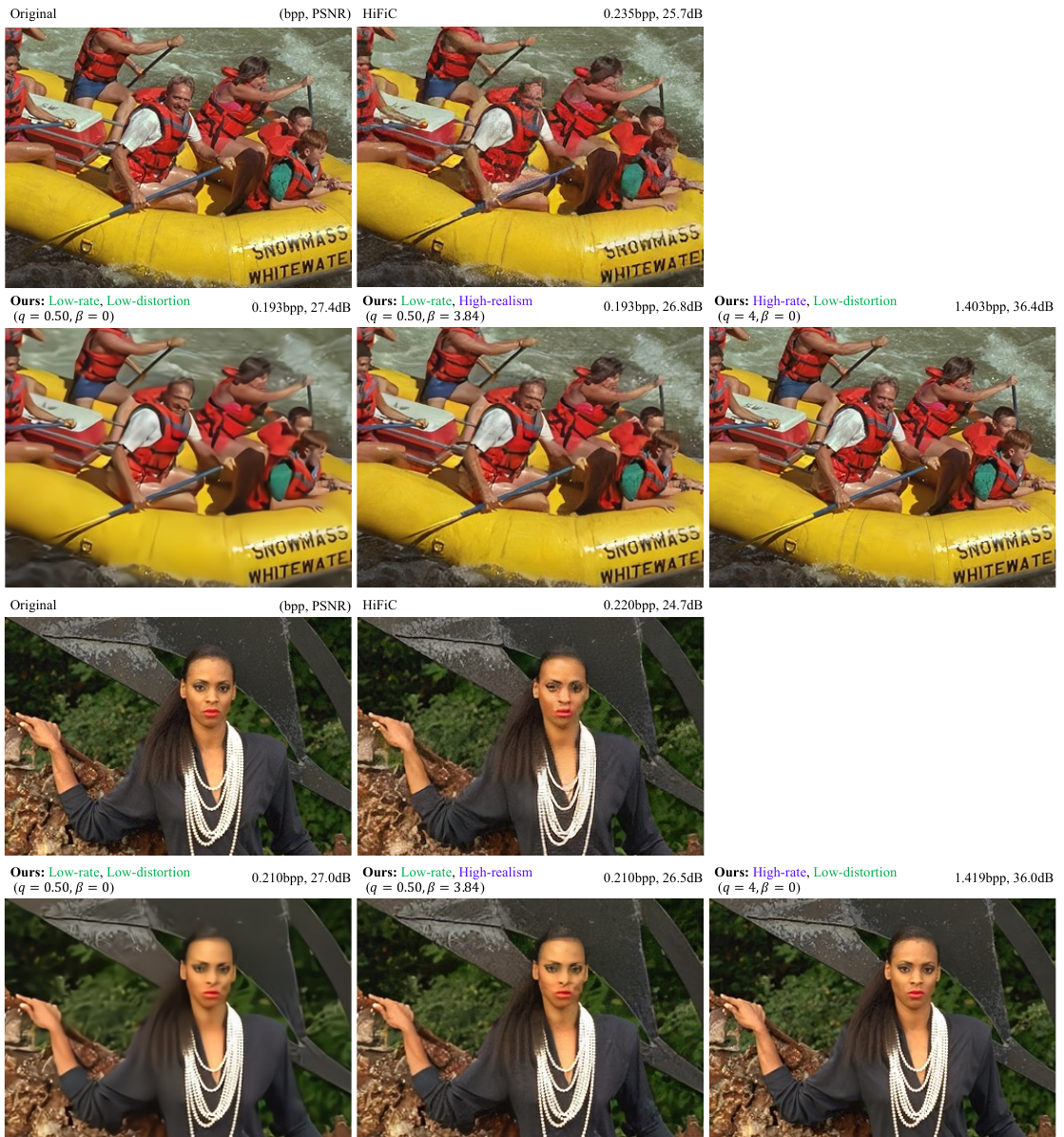}
    \caption{Qualitative comparison on Kodak dataset. In \textit{Ours}, non-integer $q$ indicates that we use interpolated channel attention\cite{Sun_2021_ACMMM} for fine rate control.}
    \label{fig:sup_kodak_qualitative}
\end{figure*}

\subsection{Additional Qualitative Results}
We show additional qualitative results in Fig~\ref{fig:sup_clic_qualitative_1},\ref{fig:sup_clic_qualitative_2},\ref{fig:sup_kodak_qualitative}. Since Kodak reconstructions of Multi-Realism\cite{Agustsson2023} are not publicly available, we compare our reconstructions with only HiFiC\cite{Mentzer2020}. Overall, our reconstructions contain fewer artifacts than HiFiC\cite{Mentzer2020} (e.g., the top figure in Fig~\ref{fig:sup_clic_qualitative_1}), and the visual quality of our method is competitive with Multi-Realism\cite{Agustsson2023}.

\end{document}